\documentclass[runningheads]{llncs}
\usepackage[T1]{fontenc}
\usepackage{graphicx}
\usepackage{subfig} 
\usepackage{makecell} 
\usepackage{cite} 
\usepackage{textgreek} 
\usepackage{amsmath} 
\graphicspath{{figures/}} 

\begin{document}
\title{Neural network-based coronary dominance classification of RCA angiograms} 
%
%
\author{Ivan Kruzhilov\inst{1, 2}\orcidID{0000-0003-1220-744X}
\and
Egor Ikryannikov\inst{3}\orcidID{0000-0002-1780-6903}
\and
Artem Shadrin\inst{4} \orcidID{0000-0001-9759-9969}
\and
Ruslan Utegenov \inst{4}\orcidID{0000-0001-8619-6478}
\and
Galina Zubkova\inst{1}\orcidID{0000-0001-9555-1689}
\and
Ivan Bessonov\inst{4}\orcidID{0000-0003-0578-5962}
}
\authorrunning{I. Kruzhilov et al.}
%
\institute{
Sber AI Lab
\email{\{iskruzhilov,gzubkova\}@sber.ru}\\
\and
Moscow Power Engineering Institute
\and
MIREA, Artificial Intelligence Institute
\and
\email{ivanbessnv@gmail.com}\\
Tyumen Cardiology Research Center, Tomsk National Research Medical Center, Russian Academy of Science\\
}

\maketitle              

\begin{abstract}
\phantom{a}
\par
\emph{Background}.
Cardiac dominance classification is essential for SYNTAX score estimation, which is a tool used to determine the complexity of coronary artery disease and guide patient selection toward optimal revascularization strategy.
\par
\textit{Objectives}.
Cardiac dominance classification algorithm based on the analysis of right coronary artery (RCA) angiograms using neural network
\par
\textit{Method}.
We employed convolutional neural network ConvNext and Swin transformer for 2D image (frames) classification, along with a majority vote for cardio angiographic view classification. An auxiliary network was also used to detect irrelevant images which were then excluded from the data set. Our data set consisted of 828 angiographic studies, 192 of them being patients with left dominance.
\par
\textit{Results}.
5-fold cross validation gave the following dominance classification metrics (p=95\%): macro recall=93.1±4.3\%, accuracy=93.5±3.8\%, macro F1=89.2±5.6\%. The most common case in which the model regularly failed was RCA occlusion, as it requires utilization of LCA information. Another cause for false prediction is a small diameter combined with poor quality cardio angiographic view. In such cases, cardiac dominance classification can be complex and may require discussion among specialists to reach an accurate conclusion.    
\par
\textit{Conclusion}.
The use of machine learning approaches to classify cardiac dominance based on RCA alone has been shown to be successful with satisfactory accuracy. However, for higher accuracy, it is necessary to utilize LCA information in the case of an occluded RCA and detect cases where there is high uncertainty. 

\end{abstract}

\section{Introduction}
The SYNTAX score, developed within the SYNTAX (Synergy Between Percutaneous Coronary Intervention with Taxus and Cardiac Surgery) study, serves the purpose of objectively evaluating the extent of severity in cases of coronary artery disease \cite{sianos2005syntax}

Currently the SYNTAX score remains as the most extensively employed and verified risk assessment tool for aiding the selection of revascularization strategies in patients with multi-vessel coronary disease \cite{lawton20222021}. Substantial limitations of this scoring system involve the intricate scoring process required for each lesion and the variation in its calculation among different physicians \cite{basman2022variability}.

\begin{figure*}[!t]
\centering
\subfloat[]{\includegraphics[height=1.6in]{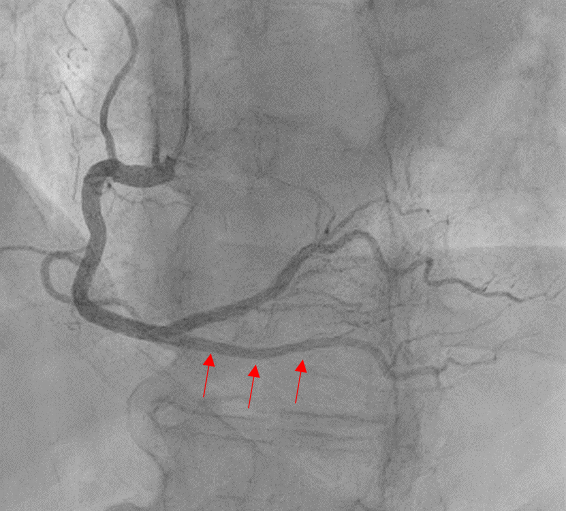}%
\label{fig_ivan_a}}
\hfil
\subfloat[]{\includegraphics[height=1.6in]{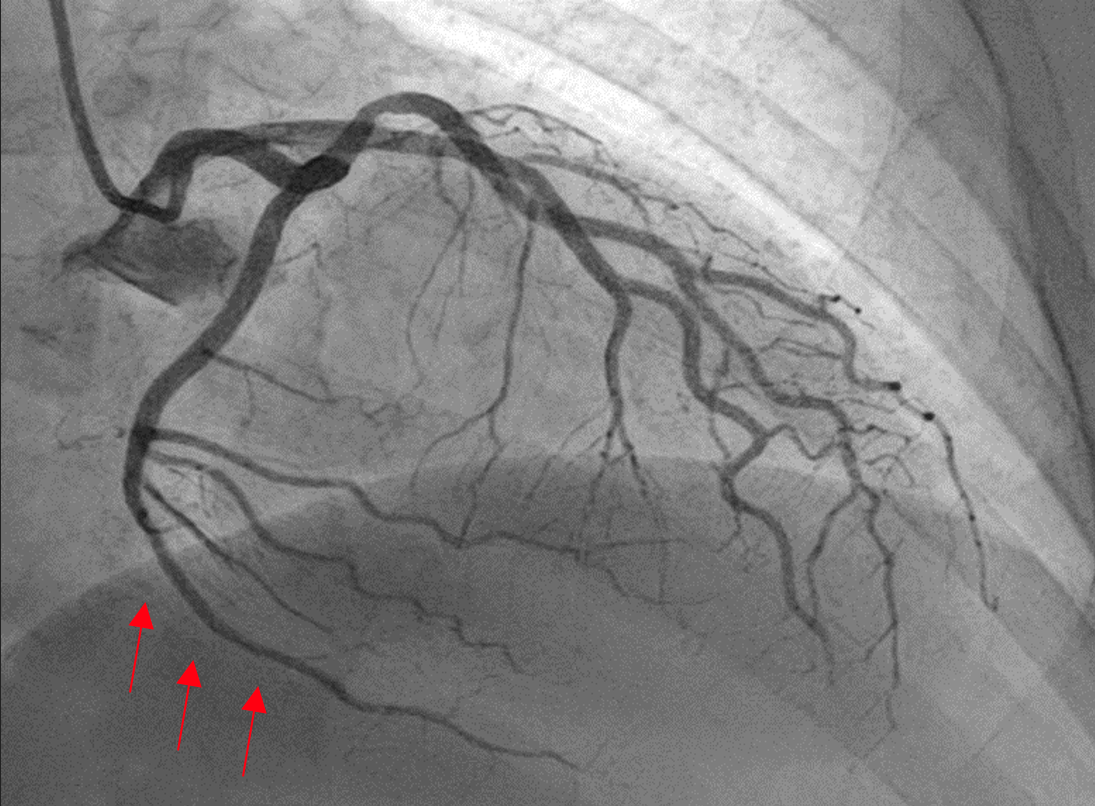}%
\label{fig_ivan_b}}
\caption{The determination of heart dominance is based on the origin of the PDA (red arrows). The PDA can originate from either the right coronary artery (a), the left coronary artery (b), or both}
\label{fig_ivan}
\end{figure*} 

The first step in calculating the SYNTAX score entails establishing cardiac dominance, a crucial stage that significantly affects the final score result. Cardiac dominance is defined by the coronary artery branch that gives rise to the posterior descending artery (PDA) (Fig \ref{fig_ivan}). It’s categorized as left, right, or co-dominant \cite{Shahoud2023dominan}. Approximately 70–80\% of the population has right dominance, where the PDA originates from the right coronary artery. About 5 to 10\% of the population shows left heart dominance, where the PDA originates from the left circumflex artery.

Furthermore, approximately 10 to 20\% of cases demonstrate co-dominance, where the PDA is supplied by both the left circumflex artery and the right coronary artery \cite{shriki2012identifying}. It is important to note that co-dominance is not considered within the options of SYNTAX score. Therefore, when dealing with these complex cases involving both arteries supplying PDA in a near equal manner; clinicians are advised to select ‘right dominance’ while calculating SYNTAX scores.


Neural networks and classical computer vision approaches \cite{cervantes2019automatic}, \cite{tao2022lightweight} can be used to automatically segment arteries, classify stenosis \cite{kaba2023application}, localize vessels and evaluate vessel diameter \cite{iyer2021angionet}. The only step remaining for complete automatic estimation of the SYNTAX score is cardiac dominance classification.

Despite progress in the right/left artery classification \cite{eschen2022classification}, the classification of cardiac dominance using neural networks needs to be solved. This may be due to the complexity of this problem, as hard samples are abundant among cardio angiographic views, which are challenging to classify. The need for a large training data set with balanced classes (as left dominant hearts usually represent only 10\% or less) further complicates matters. 

Our research is limited to Right coronary artery (RCA) analysis. This study provides a baseline for future improvements in this field and highlights the complexity of cardiac dominance classification tasks, making it a valuable contribution to both cardiology and machine learning communities. 

\paragraph{ This article presents the fully automatic cardiac dominance classification using neural networks.} The \textit{contributions} of the article are:
\begin{itemize}
\item We have successfully collected and annotated a sizeable angiographic study data set, enabling us to classify cardiac dominance accurately.
\item Our research has demonstrated the superiority of symmetrical and normalized cross-entropy over conventional cross-entropy regarding accuracy and ROC AUC when dealing with noisy labeled data such as bad-quality frames, specific anatomy, or artifacts.
\item We have used an ensemble of models to identify subsets of angiographic studies where ML models fail due to specific anatomy; these cases must be treated specially for accurate predictions.
\end{itemize}

\section{Methodology}
The approach for the cardiac dominance classification is as follows. We analyze each angiogram projection frame-wise, and only those deemed informative frames take part in the decision-making process. A specially trained neural network is used to determine which frames are considered informative or not. The final decision for an angiogram view is then made by a majority voting based on the results of all relevant frames analyzed. The majority voting is an unweighted sum of all informative frames logits.

\begin{figure}[t] 
\centering
\includegraphics[width=0.85\textwidth]{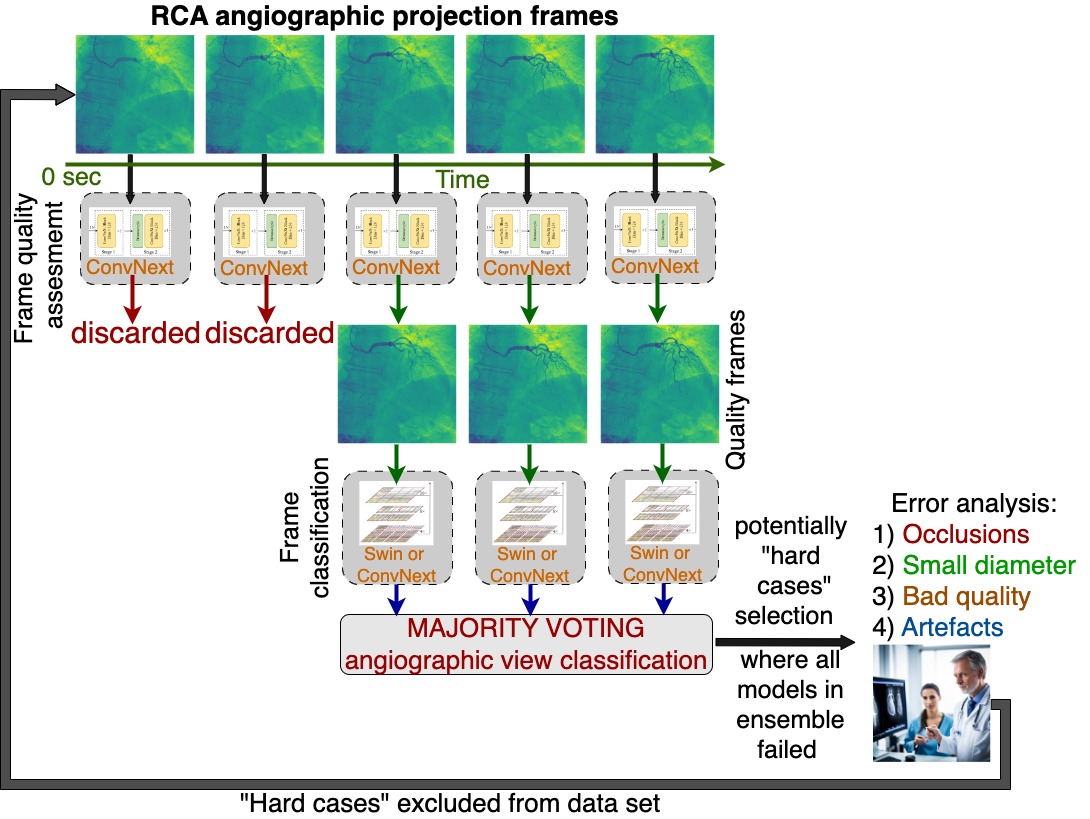}%
\caption{Proposed coronary dominance classification pipeline. }
\label{fig_segment_scheme}
\end{figure} 

Fig. \ref{fig_segment_scheme} demonstrates the scheme of our classification algorithm and experimental design. To evaluate its performance, we employed a 5-fold cross-validation approach.

Furthermore, an ensemble of neural networks was used to identify hard cases for which the model stumbled over during classification. Those studies that all models in the ensemble failed to classify correctly were labeled with additional tags clarifying why they had been misclassified. To estimate the influence of hard cases on the classification metrics, we excluded them from the data set and trained the model a second time

\label{section_dataset}
\subsection{Coronary angiogram dataset}
This retrospective study included 850 patients who underwent invasive coronary angiography between January 2021 and August 2022 in the Tyumen Cardiology Research Center – Branch of Tomsk National Research Medical Center (Russian Federation). The study complies with the principles of the Declaration of Helsinki. The study protocol was approved by the local ethics committee.

The invasive coronary angiography procedures were performed following the latest guidelines, utilizing an interventional angiography system (Phillips Allura Clarity, USA), and were acquired at 15 frames/sec.

The data set contains 828 angiographic studies. Out of 850 initial studies included in the data set, 22 (2.6\%) had low image quality or contrasting conditions that led them being excluded from training; thus leaving 192 left dominant cases out of 828 remaining ones and 636 right dominant cases.

This data indicates that the average patient is 63.6 years old, with 17.6\% of patients being younger than 55, 35.5\% in the range of 55-65, 36.1\% in the range 65-75, and 10.9\% above 75 years old, respectively; additionally 59\% are male and 41\% female patients overall. 

Coronary angiographic (CA) view (projection) is a tool for medical imaging, allowing scientists to observe the body from a 3D space in real-time and capture it on 2D imaging planes. This technique involves X-ray technology to create video images of internal organs and vessels. The study consists of LCA and RCA views, with 1-3 RCA views and 1-2 views for the LCA on average. A CA view comprises an average of 30-60 frames, where pixels are linearly normalized to a range from 0 to 255.

The data set contains 1465 CA views, with 235 for the left and 1230 for the right type. Despite there being more patients with left dominant hearts in this data set than on average in the population, it still suffers from an imbalance of one to five due to a higher number of RCA views for right dominant studies (2.0) compared to those found in left dominant studies (1.2). 

Three physicians with 3-15 years of experience took part in the data labeling process. Each physician was assigned one chunk of data, and each study was labeled only once by a single physician

\subsection {Frame selection}
\label{section_dataset}
Using a contrast medium is essential for x-ray imaging, as it allows the vessels to be seen on an x-ray. However, because contrast medium spreads over time in the vessels and attenuates during its course, this can lead to a need for selecting key frames in the input video~\cite{kharchevnikova2021efficient}.

CathAI \cite{avram2021cathai} used an algorithm for automatically identifying peak-contrast frames in angiographic videos. The method calculates the structural similarity index between each video's first "reference" frame (which typically does not contain intra-arterial contrast) and all subsequent frames, then selects the frame with the lowest similarity index as most likely containing peak contrast. Up to 8 frames can be extracted from each video using this approach.

The disadvantage of this method is that the number and quality of frames used in studies can vary significantly, leading to some CA views being unsuitable for model training due to lack of high quality. As such, ensuring that only those studies with sufficient high-quality data are included in the model training process is essential.

We have undertaken another approach for the annotation of informative and non-informative frames. We manually annotated 253 CA views (114 left type, 139 right type) by setting the range of informative frames with a margin of 5 to categorize any frame outside this range as non-informative. This process resulted in 7713 (informative 4218, non-informative 3495) slices being labeled without involving radiologists, and our criteria for selecting informative frames was based on clear visibility of artery’s branches.

We trained a neural network (ConvNext tiny) to label the remaining frames and used it to exclude non-informative frames from the training. We found empirically that different image casting thresholds should be used for different dominance types; namely, the threshold for right type should be higher (0.6) than for the left one (0.3); this is due to RCA being thicker in right dominant hearts and also because of left dominant hearts being rarer in our data set. This type of thresholding is only applicable during the training process as cardiac dominance can then be known beforehand; however, when testing images with this model, we use a single unified threshold value of 0.55 across all images regardless of their respective dominance types.

\begin{figure*}[!t]
\centering
\subfloat[]{\includegraphics[width=1.2in]{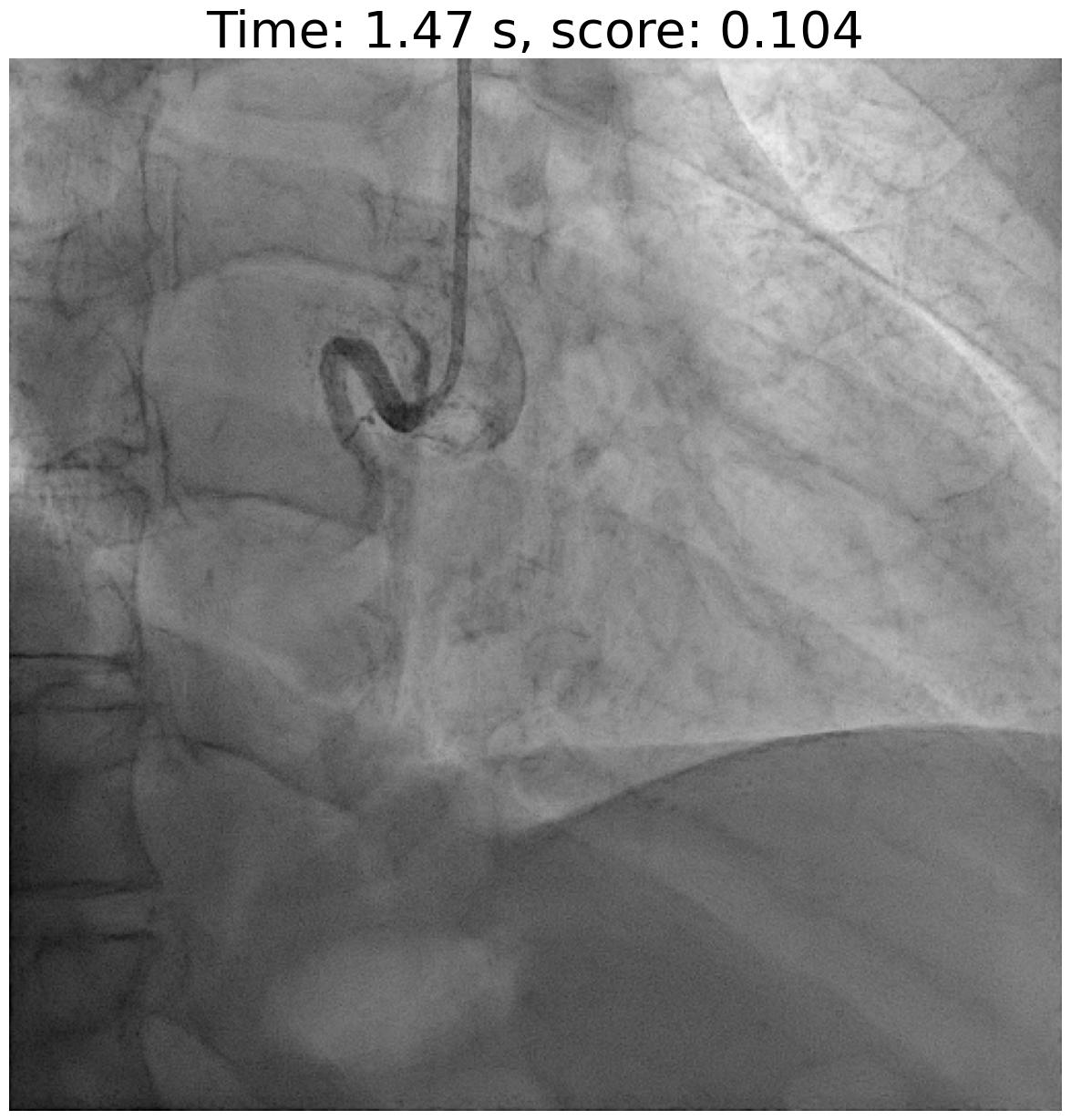}%
\label{fig_first_case}}
\hfil
\subfloat[]{\includegraphics[width=1.2in]{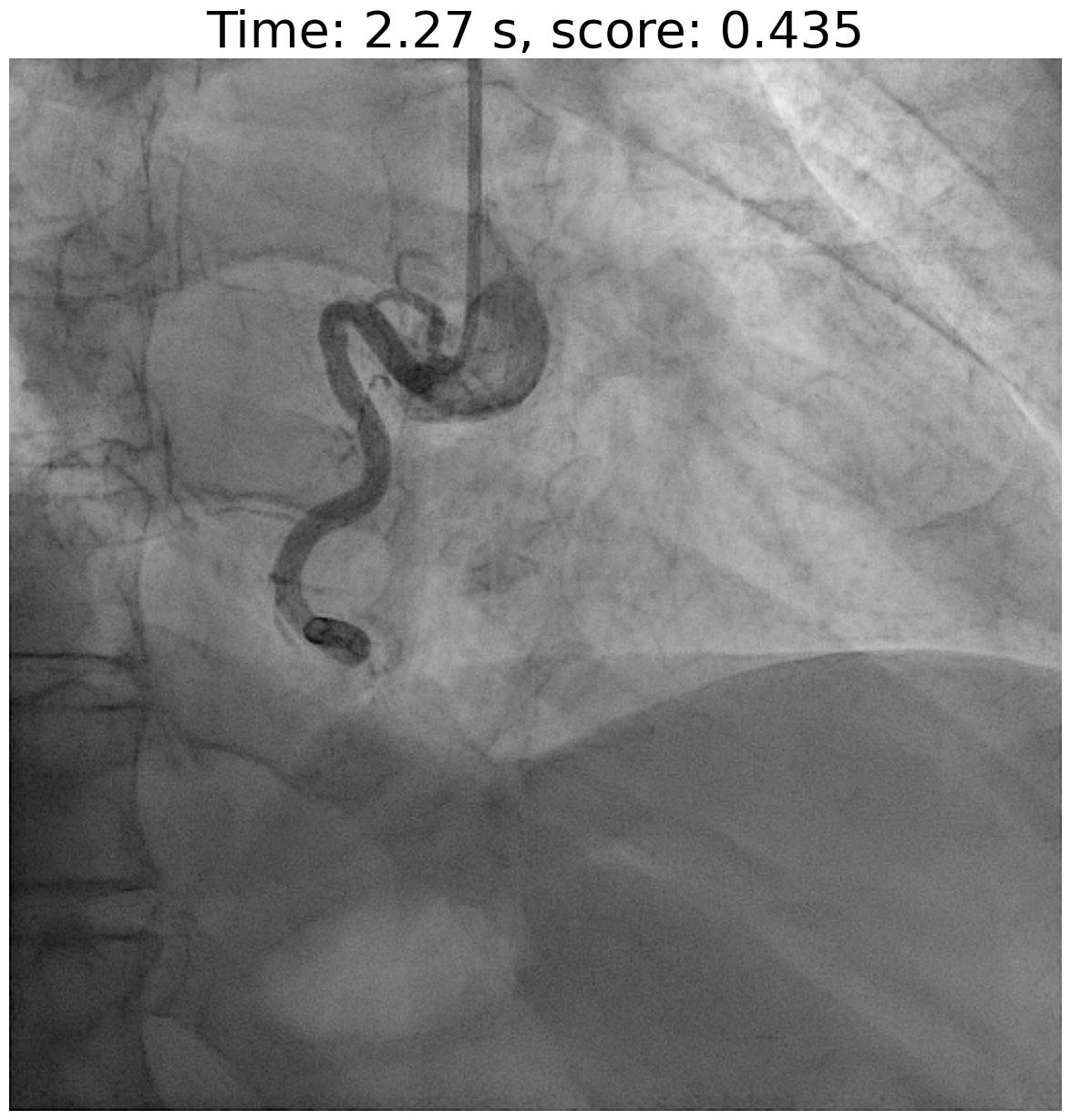}%
\label{fig_second_case}}
\hfil
\subfloat[]{\includegraphics[width=1.2in]{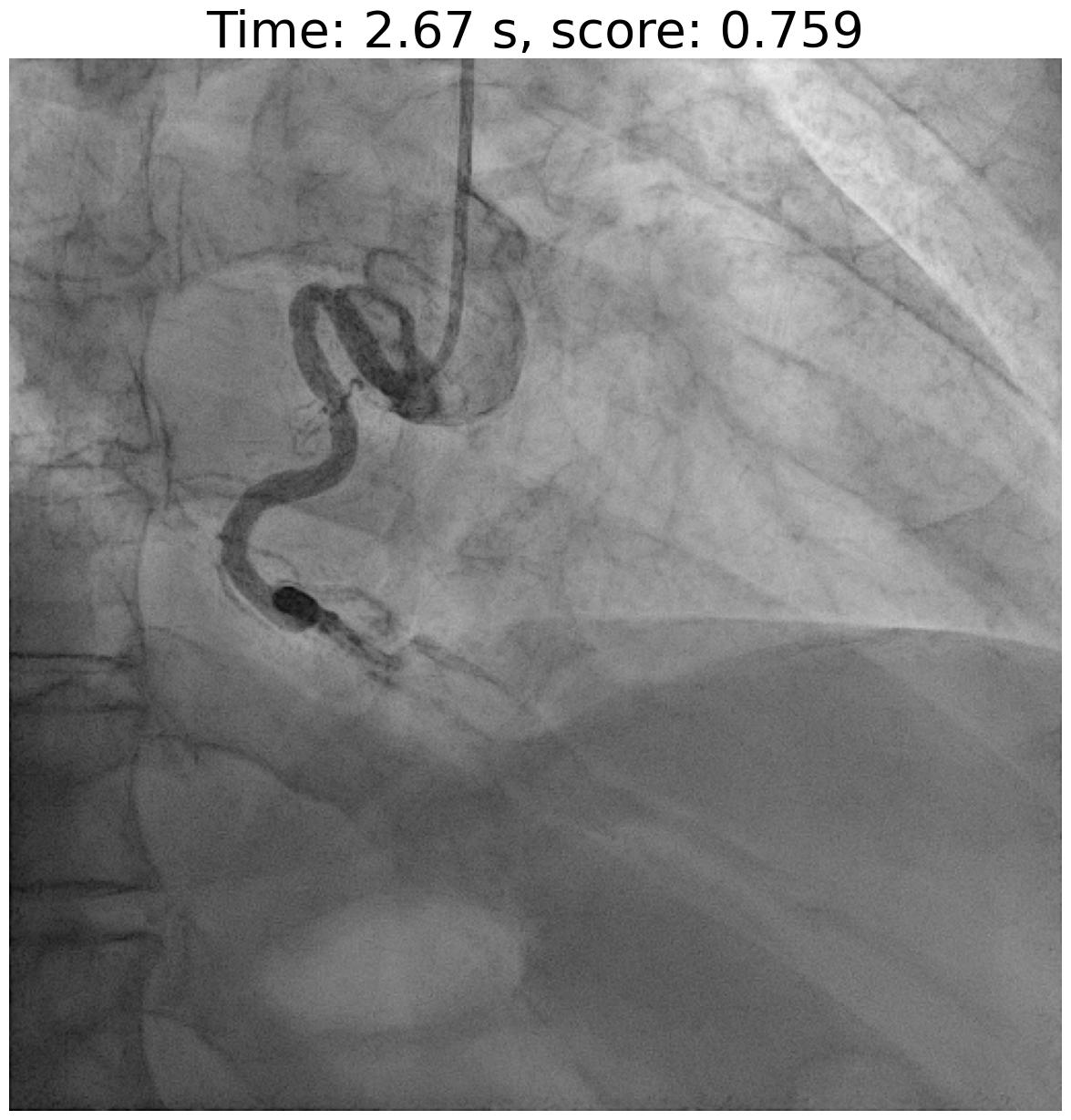}%
\label{fig_third_case}}
\hfil
\subfloat[]{\includegraphics[width=1.2in]{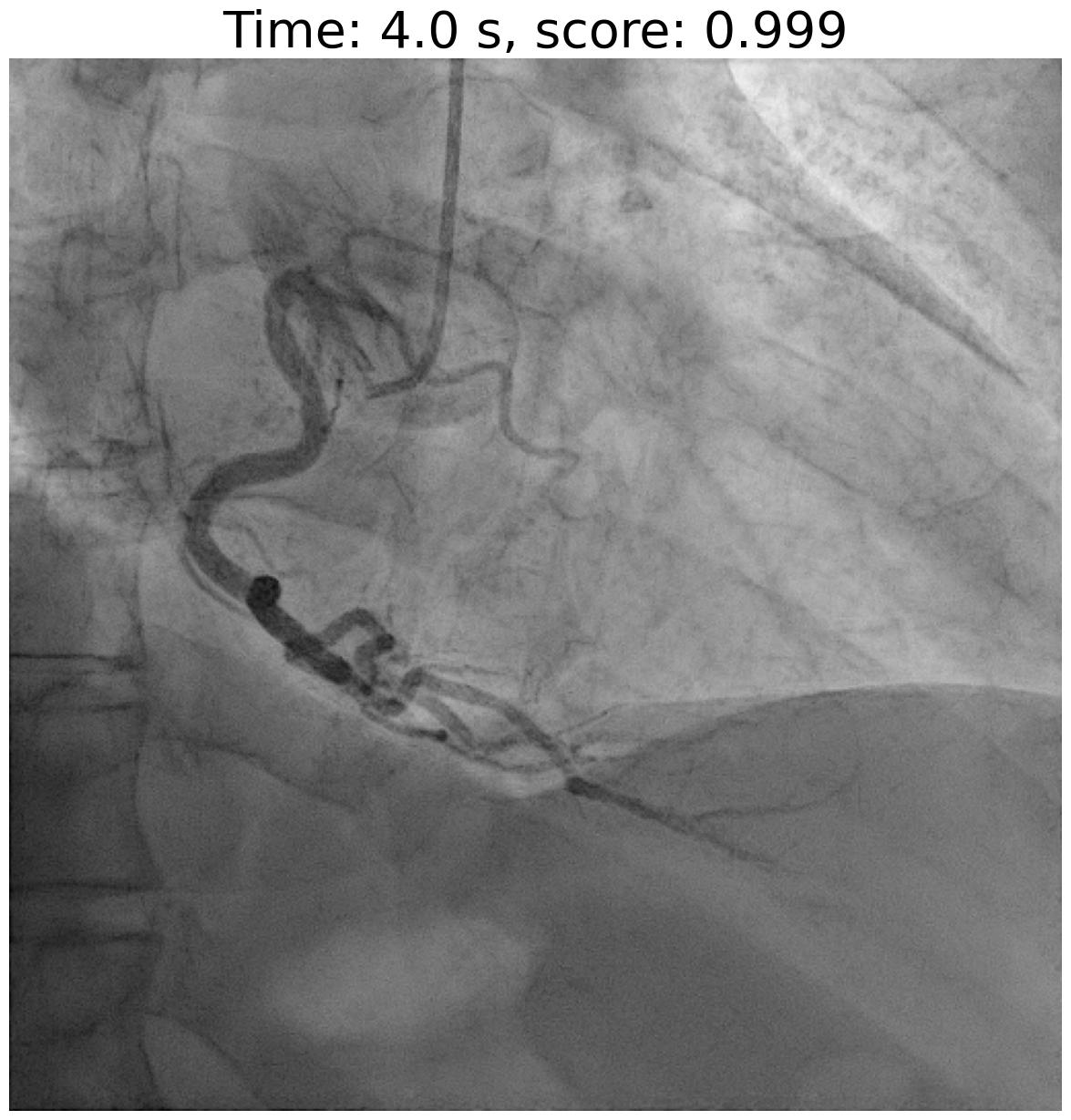}%
\label{fig_fourth_case}}
\caption{RCA frame quality score. 0 - low quality, 1 - high qualitye, (a) 1.47 s, score=0.104. (b) 2.27 s, score=0.435 (c) 2.67 s, score=0.759 (d) 4.00 s, score=0.999}
\label{fig1_quality_estim}
\end{figure*}

Fig. \ref{fig1_quality_estim}  demonstrates how a neural network estimates frame quality score for RCA over time. The total time of the angiogram is 4870 ms.

\subsection{Neural network implementation and training details}
\label{section:neural_network}
We have trained two distinct models -  a convolutional neural network ConvNext \cite{liu2022convnet}, and Swin \cite{liu2021swin}, a hierarchical vision transformer using shifted windows. These models achieved the best results on multiple benchmarks across various tasks, such as image classification and object detection. Despite their similar size (Table \ref{table:model_parameters}), their architecture differs significantly; while CNNs are based on a series of layers that apply filters to an input image, Transformers use attention-based methods to recognize patterns within images. 

When training these models with pretrained weights from ImageNet datasets, we can expect better performance than when randomly initialized weights are used instead. To take full advantage of this transfer learning approach, we resized the original 512*512 images down to 224*224 resolution and converted grayscale inputs into 3 channels so they match those found in ImageNet datasets more closely. This way our model will be able to benefit from already learned features without needing additional time or resources during  training process, which would otherwise be necessary if random initialization was used instead.

\begin{table}[!t]
\caption{Model and training parameters. \label{tab:table4}}
\label{table:model_parameters}
\centering
\begin{tabular}{|c||c||c||c||c|}
\hline
\thead{Method} & \thead{parameters \\ mil.} &\thead{batch size} &\thead{epochs } &\thead{Min - max \\ lr}\\
\hline \makecell{ConvNext} & 27.8 &100 &7 &10$\textsuperscript{-5}$ - 0.0001\\
\hline \makecell{Swin} &27.6 &100 &7 &10$\textsuperscript{-5}$ - 0.0001\\
\hline
\end{tabular}
\end{table} 

\paragraph{Loss function.}

We used a normalized symmetrical cross-entropy (NSCE) loss function (\ref{equation:NSCE}) to mitigate noisy labelling \cite{ma2020normalized}.
\begin{equation}
\label{equation:NSCE}
NSCE = \alpha NCE + \beta RCE
\end{equation}

This upgraded version (\ref{equation:NCE}) of the original normalized cross-entropy (NCE) definition incorporates a class weight $w_k$, as well as ground truth probability, $q(k|x)$, and estimated probability, $p(k|x)$, for each class $k$ in $K$ classes. We have defined $q_j$ to be a one-hot vector with one located at j position and $\omega _{norm}=\sum_{k=1}^{K}\omega _k q(k|x)$ is the normalization coefficient. The weighted form of NCE function helps us better manage imbalanced data sets while training deep learning models.

\begin{equation}
\label{equation:NCE}
NCE = \frac{1}{\omega _{norm}}\frac{\sum_{k=1}^{K}\omega _k q(k|x)log(p(k|x))}
{\sum_{k=1}^{K}{\sum_{j=1}^{K}{q_j(k|x)log(p(k|x))}}}
\end{equation}

Reverse cross-entropy (RCE) \cite{wang2019symmetric} is a noise tolerant term:
\begin{equation}
RCE = \frac{1}{\omega _{norm}}\sum_{k=1}^{K}\omega _kp(k|x)log(q(k|x))
\end{equation}
where $q(k|x)$ = 0 is to be substitute with a small positive value. We used $\alpha$ = 0.3 and $\beta$ = 0.7 in all experiments.

\paragraph{Training details.}
During the training, we used only every second frame for the right dominant class to overcome the class imbalance.  We also exploited weighted normalized symmetrical cross-entropy for this purpose with weights 0.28 and 0.72 for the right and left dominant classes.
To reduce the overfitting, we applied the following augmentations – 15 deg rotation and crop (scale=(0.8, 1), ratio=(0.85, 1.15)).  

We used the standard Torchvision implementations of ConvNext and Swin models with Torch 2.0.1+cu117 and Torchvision 0.15.2 running on an NVIDIA V100 GPU for training purposes, employing AdamW as the optimizer with a weight decay of 0.05. We trained ConvNext with a constant lr=0.0001 after zero epoch and lr=10\textsuperscript{-5} for zero epoch. An appropriate learning schedule was selected for Swin training starting from a warmup period at a low initial learning rate (lr) of 10\textsuperscript{-5} up to its maximum lr value set at 0.0001 and then decreased to 0.00005 after the 4\textsuperscript{th} epoch.

\section{Results}
\label{section_results}

\subsection{Metrics}
\label{section:metrics}
The four main metrics we use to evaluate our models are $recall_{macro}$, $accuracy$, $F1_{macro}$, and $MCC$ (Matthew correlation). These scores can be calculated by taking into account True Positive (TP), False Positive (FP), True Negative (TN) and False Negative (FN). We also calculate a separate score for right-dominance classifications called $recall_{right}$ or $precision_{right}$, where positive outcomes refer to right-dominance while negative outcomes refer to left-dominance; similarly, we have a separate score for left-dominance called $recall_{left}$ and $precision_{left}$.
\begin{equation}
\label{equation:recall}
recall =\frac{TP}{TP + FN}
\end{equation}

\begin{equation}
\label{presicion}
precision =\frac{TP}{TP + FP}
\end{equation}

The $F1$ score (\ref{equation:f1}) is a useful metric for evaluating the performance of a model, however it has one major disadvantage: it does not consider the number of correctly classified negative samples. The $F1_{macro}$ solves (\ref{equation:f1_macro}) this problem by taking into account both true positives and true negatives in its calculation.

\begin{equation}
\label{equation:f1}
F1 =2\frac{precision*recall}{precision + recall} = \frac{2TP}{2TP + FP + FN}
\end{equation}

\begin{equation}
\label{equation:f1_macro}
F1_{macro} = \frac{F1_{left} + F1_{right}}{2}
\end{equation}

A popular alternative for $F1_{macro}$ is $MCC$ (\ref{equation:MCC}). $MCC$ \cite{chicco2020advantages} is also invariant for class swapping and therefore is a macro metrics for binary classification. 

\begin{equation}
\label{equation:MCC}
MCC = \frac{TP*TN - FP*FN}{\sqrt{(TP+FP)(TP+FN)(TN+FP)((TN+FN)}}
\end{equation}

The precision and correlated with it $F1$ and $MCC$ metrics are sensitive to the class imbalance in the test data.  For example, suppose there is an unequal distribution of samples between classes, such as 10 samples of class 1 and 90 samples of class 2. In that case, a model that performs equally well for both classes may still have a low precision score for class 1 due to false predictions. $Accuracy$ (\ref{equation:accuracy}) is not sensitive enough to this issue either since it does not account for false positives from rarer labels. $Recall_{macro}$ (\ref{equation:recall_macro}), however has been shown to be insensitive to imbalanced data sets and thus provides a more accurate assessment when cross-validating models on unbalanced data sets.

\begin{equation}
\label{equation:accuracy}
accuracy =\frac{TP + TN}{TP + TN + FP + FN}
\end{equation}

\begin{equation}
\label{equation:recall_macro}
recall_{macro} = \frac{recall_{left} + recall_{right}}{2}
\end{equation}

MCC, F1, accuracy, precision, and recall are performance assessment measures for CA views classification. In addition to these performance assessment measures, ROC AUC (Receiver Operating Characteristic - Area Under the Curve) is also helpful when assessing individual frame classification as it measures how well the model distinguishes between classes at various thresholds.

\subsection{Experiment design and classification results}
\label{section:experiment_design}
We used a 5-fold cross-validation strategy, splitting the studies into five equal folds to avoid data leakage. However, the number of projections and slices may differ between folds due to varying amounts for different studies. To ensure accurate results were achieved during training, we reserved 20\% of each fold's train part for validation (16\% of the total data set). This train/validation split was not fixed like a 5-fold test/train split. Therefore, we regulated this split by fixing random seed in order to compare results correctly between experiments conducted on different folds and splits. In total, four separate 5-fold splits with two distinct train/validation splits were completed resulting in twenty neural networks trained overall.


Table \ref{table:recall_precision} and Table \ref{table:macro_metrics} provide a comprehensive view of the performance of right and left dominance classification for a CA view and individual frame classifications. The data featured in both tables include recall, precision, ROC AUC metrics for individual frames (Table \ref{table:recall_precision}) along with macro metrics such as accuracy, recall macro, F1 macro, and MCC (Table \ref{table:macro_metrics}). All margin intervals ($t_{st}\sigma$) are estimated with 95\% confidence based on statistics from 20 different neural networks trained through four different 5-fold cross-validation.

\begin{table}[!t]
\caption{Cardiac dominance classification. Recall and precision for a CA view classification.}
\label{table:recall_precision}
\centering
\begin{tabular}{|c||c||c||c||c||c|}
\hline
\thead{Method} & \thead{Recall \\ left $\uparrow$} &\thead{Recall \\ right $\uparrow$} &\thead{Precision \\ left $\uparrow$} &\thead{Precision \\ right $\uparrow$} &\thead{ROC AUC $\uparrow$ \\ frame class}\\
\hline \makecell{ConvNext} & 92.4$_{\pm9.3}$\% &\textbf{93.8}$_{\pm5.0}$\% &\textbf{74.6}$_{\pm14.8}$\% &98.5$_{\pm1.8}$\% &\textbf{95.7}$_{\pm4.5}$\%\\ 
\hline \makecell{Swin} & \textbf{94.9}$_{\pm9.7}$\% &91.4$_{\pm5.4}$\% &68.5$_{\pm12.2}$\% &\textbf{99.0}$_{\pm1.9}$\% &93.1$_{\pm5.4}$\%\\
\hline
\end{tabular}
\end{table} 

\begin{table}[!t]
\caption{Cardiac dominance classification. Accuracy, macro recall, F1, and MCC for a CA view classification.}
\label{table:macro_metrics}
\centering
\begin{tabular}{|c||c||c||c||c|}
\hline
\thead{Method} & \thead{Recall \\macro $\uparrow$} &\thead{Accuracy $\uparrow$} &\thead{F1 macro $\uparrow$} &\thead{MCC $\uparrow$}\\
\hline \makecell{ConvNext} & 93.1$_{\pm4.3}$\% &\textbf{93.5}$_{\pm3.8}$\% &\textbf{89.2}$_{\pm5.6}$\% &\textbf{79.3}$_{\pm10.1}$\% \\ 
\hline \makecell{Swin} &\textbf{93.2}$_{\pm4.5}$\% &92.0$_{\pm4.2}$\% &87.2$_{\pm5.5}$\% &76.3$_{\pm9.6}$\% \\
\hline
\end{tabular}
\end{table} 

\subsection{Analysis of the cases where neural network failed}

The neural network failed to classify, on average, about 70 out of 828 studies, so we employed an ensemble of five models to search for studies with hard cases or potentially wrong labeling. We hypothesized that if all models failed to classify a study correctly, this was more likely due to data-related issues rather than model-related ones. The ensemble consisted of ConvNext and Swin models with different fold splits. After searching through 41 such studies (5 left and 36 right dominant) containing 51 CA views, they were further labeled by a highly skilled cardiac surgeon who holds a PhD and has over 15 years of experience in the field.

All cases are considered to be hard. The second labelling showed that only 10\% of examined studies were incorrectly labeled due to atypical branching. The most frequent case (61\%) in which the model failed was RCA occlusion, as it is almost impossible to accurately classify dominance without analyzing the LCA and RCA. Consequently, our model trained on the RCA alone could not distinguish this case.

\begin{figure*}[!t]
\centering
\subfloat[]{\includegraphics[height=1.6in]{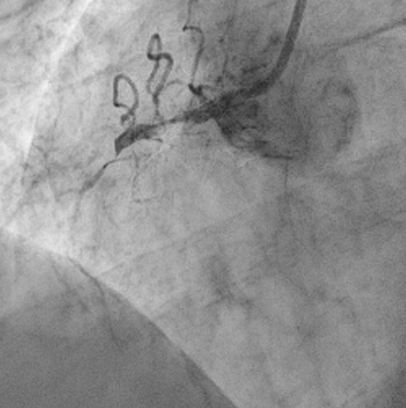}%
\label{fig_RCA_occlus}}
\hfil
\subfloat[]{\includegraphics[height=1.6in]{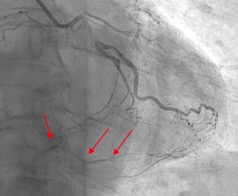}%
\label{fig_LCA_collat}}
\label{fig_occlusion}
\caption{An example of occlusion in RCA. Red arrows point to the collateral channels. (a) RCA, (b) LCA.}
\end{figure*}

Fig. \ref{fig_RCA_occlus} illustrates an occlusion in the RCA, making it impossible to determine dominance using only this frame. However, Fig. \ref{fig_LCA_collat} shows a clearly visible contrast medium in collateral channels, which indicates that the model has misclassified a right dominant heart as the left dominant one. This example highlights the importance of utilizing RCA and LCA frames when determining cardiac dominance.

\begin{figure}[t] 
\centering
\includegraphics[width=2.0in]{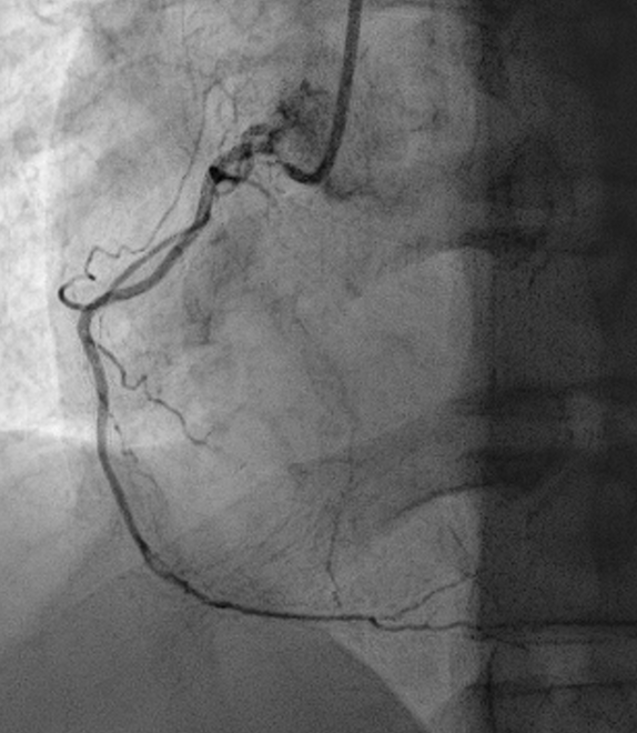}%
\caption{An example of small right coronary artery complicating dominance classification.}
\label{fig_athypical_branch}
\end{figure}

Table \ref{table:false_labeled} provides an overview of 41 cases where the model ensemble classified incorrectly. The first row indicates the percentage of these cases among all instances, while rows two and three reveal the proportion of impossible-to-predict correct results without an LCA analysis, as well as the ratio between false labels. Fig. \ref{fig_athypical_branch} illustrates the second popular reason for false classification - a small right coronary artery.

Only 5\% of cases had low-quality angiograms, indicating that our system successfully identified and addressed this issue as the most common cause of incorrect dominance classification. This finding demonstrates that the frame quality scoring system we have implemented effectively prevents errors due to poor CA views.

The rest 2.4\% of the errors are due to the artefacts -  electrode catheters for the cardiac stimulation.

\begin{table}[!t]
\caption{Summary of the cases where the all models in ensemble failed}
\label{table:false_labeled}
\centering
\begin{tabular}{|c||c||c||c||c|}
\hline
\thead{Cases} & \thead{Occlusion} &\thead{small \\ artery} &\thead{Low \\ quality} &\thead{Artefacts}\\
\hline \makecell{\% of total} & 61.0\% &31.7\% &4.9\% &2.4\% \\ 
\hline \makecell{require LCA} &88\% &23.1\% &0\% &0\% \\ 
\hline \makecell{false labeled} &0\% &30.8\% &0\% &0\% \\ 
\hline
\end{tabular}
\end{table} 

\subsection{The sufficiency of the data set size}

In order to demonstrate that the model stumbles not only over hard cases but the limitation of the data set size as well, we conducted an experiment in which we trained the ConvNext model on different subsets of our main data set containing 20\%, 25\%, 30\%, 40\%, 50\% and 75\% of its size. We then excluded 41 hard cases (5\% of the data) from this reduced data set and retrained our model with it.

\begin{figure}[t] 
\centering
\includegraphics[width=2.7in]{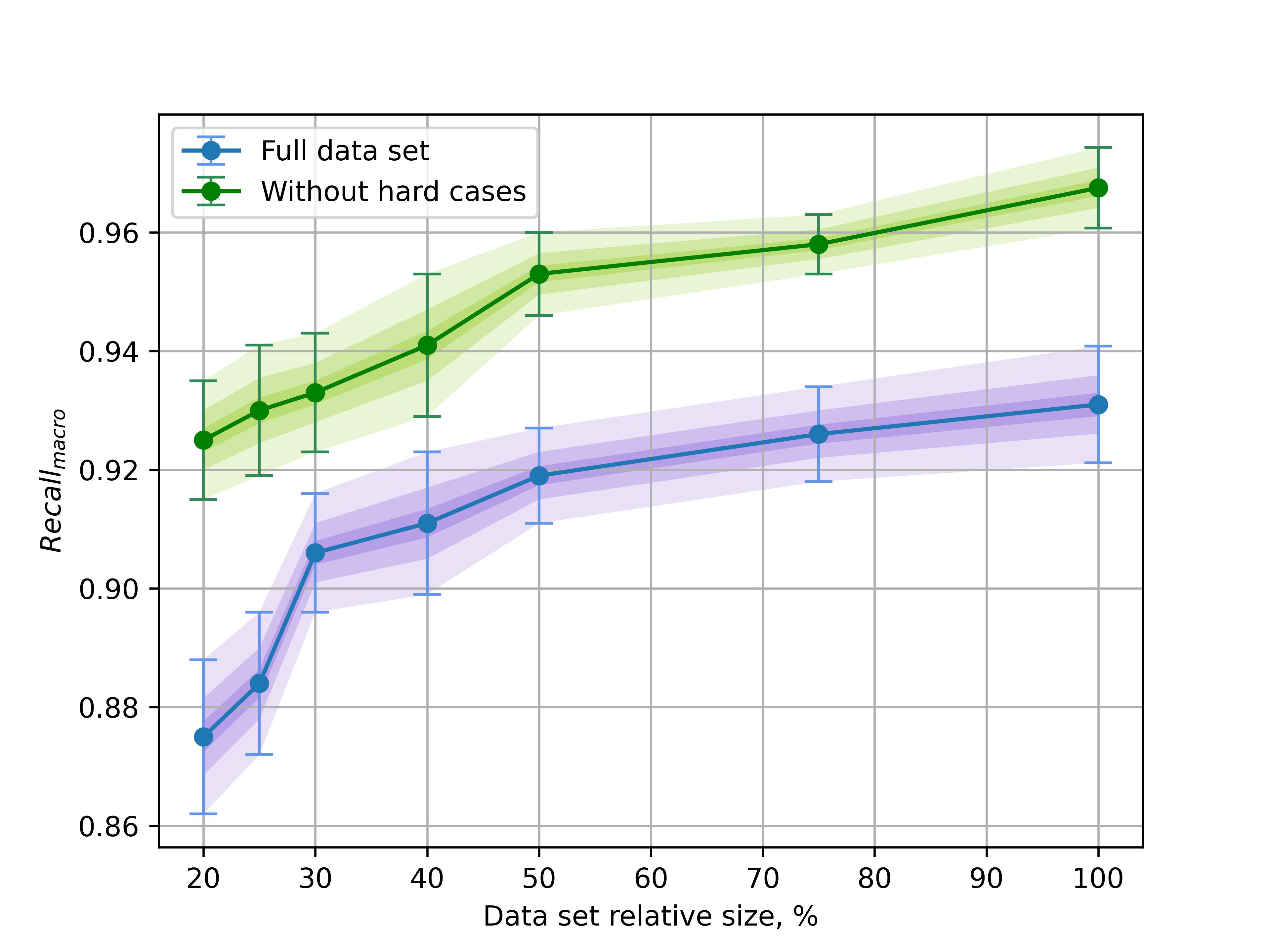}
\caption{Dependence of mean macro recall on relative data set size. 95\% confidence interval. Full data set (blue line) and data set without 41 hard cases (green line) }
\label{fig_saturation}
\end{figure}

Fig. \ref{fig_saturation} demonstrates the dependence of $recall_{macro}$ on the relative data set size, with two lines representing the mean value and 95\% confidence intervals ($t_{st}\sigma / \sqrt{n}$). It is important to note that these confidence intervals should not be confused with those found in Tables 1-2. We conducted multiple experiments to estimate std for various 5-fold validations of different shuffles (from 28 for 20\% of data to 4 for 100\%). 

The steady increasing trend of the curves suggests that further improvements in accuracy may be possible by adding additional data to our data set.

\subsection{Ablation study}
We have conducted a comparison between our method and frame selection based on inter-frame SSIM comparison \cite{avram2021cathai}. For training purposes, we considered a range of four frames surrounding the peak frame, which exhibited the highest diversity in SSIM when compared to the initial frame. The findings presented in Table \ref{table:frame_selection} indicate that this particular approach yielded less favorable classification metrics when utilizing all frames for classification.

If we do not use our additional frame selection, the classification accuracy and ROC AUC degrades from 93.6\%  and 95.7\%  to 90.9\% and 89.8\% , respectively. Discarding the first twenty frames as bad quality is a reasonable approach, as the contrast medium needs to spread more at the angiography's beginning. It is better than using all frames for classification but does not outperform our method. 

\begin{table}[!t]
\caption{Cardiac dominance classification by ConvNext with and without frames selection based on their quality. 95\% confidence interval for mean value.}
\label{table:frame_selection}
\centering
\begin{tabular}{|c||c||c||c||c|}
\hline
\thead{Method} & \thead{Recall \\macro $\uparrow$} &\thead{Accuracy $\uparrow$} &\thead{F1 macro $\uparrow$} &\thead{ROC AUC $\uparrow$}\\
\hline \makecell{frame \\ selection \\ (our)} & \textbf{93.1}$_{\pm1.0}$\% &\textbf{93.6}$_{\pm1.0}$\% &\textbf{89.1}$_{\pm1.2}$\% &\textbf{95.7}$_{\pm1.0}$\% \\
\hline \makecell{frame \\ selection \\ SSIM \cite{avram2021cathai}} &88.5$_{\pm0.9}$\% &90.4$_{\pm1.2}$\% &84.2$_{\pm1.3}$\% &94.4$_{\pm0.8}$\% \\

\hline \makecell{first 20 \\frames \\discarding} &90.6$_{\pm1.2}$\% &88.6$_{\pm2.2}$\% &82.9$_{\pm2.5}$\% &94.9$_{\pm0.9}$\% \\

\hline \makecell{all frames} &88.5$_{\pm1.1}$\% &90.9$_{\pm0.9}$\% &84.8$_{\pm1.3}$\% &89.8$_{\pm0.8}$\% \\
\hline
\end{tabular}
\end{table} 

Our experiments have shown that the random flips, mix-up \cite{zhang2017mixup}, RandAUG \cite{cubuk2020randaugment}, optical and grid distortion augmentations used along with rotation and crop reduced overfitting, but dropped the metrics or at least did not improve them. 

Cross-entropy (CE) loss is widely used for classification problems dbecause it minimizes the KL-divergence between two discrete distributions. However, it suffers from the same drawbacks as KL-divergence and can be vulnerable to outliers \cite{ma2020normalized}. An alternative approach is Symmetrical cross-entropy (SCE), which belongs to the class of active/passive losses and has been shown in studies such as \cite{wang2019symmetric} to be more stable when dealing with corrupted labels.

Our study compared conventional CE, SCE, and NSCE for cardiac dominance type classification. We used for SCE the same $\alpha$ = 0.3 and $\beta$ = 0.7 as for NSCE.

The results showed that NSCE provided the best result (Tables \ref{table:loss_comparison}, \ref{table:loss_comparison_swin}). Specifically, the ROC AUC in Tables \ref{table:loss_comparison} and \ref{table:loss_comparison_swin} indicates frame classification accuracy, while recall macro stands for CA view classification performance. The confidence interval is for the mean value - $t_{st}\sigma / \sqrt{n}$. These findings suggest that NSCE and SCE are viable options when classifying cardiac dominance and  perform significantly better over conventional cross-entropy.

\begin{table}[!t]
\caption{ConvNext classification metrics for different loss functions. 95\% confidence interval for the mean value}
\label{table:loss_comparison}
\centering
\begin{tabular}{|c||c||c||c||c||c||c|}
\hline Model & \multicolumn{3}{c}{\thead{ROC AUC $\uparrow$ \\ frame \\ classification}} & \multicolumn{3}{c|}{\thead{Recall macro $\uparrow$ \\ CA view \\classification}} \\
\hline
&\thead{40\% data} & \thead{75\% data} & \thead{100\% data} &\thead{40\% data} &\thead{75\% data} &\thead{100\% data}\\
\hline \makecell{CE} &$92.8_{\pm0.7}$\% &$94.7_{\pm0.5}$\% &$95.2_{\pm0.81}$\% &$86.9_{\pm1.3}$\% &$90.2_{\pm0.8}$\% &$91.1_{\pm1.3}$\%\\
\hline \makecell{SCE=CE+RCE} &\textbf{94.7}$_{\pm0.6}$\% &\textbf{95.7}$_{\pm0.5}$\% &\textbf{96.1}$_{\pm0.7}$\% &90.6$_{\pm1.0}$\% &$92.2_{\pm0.8}$\% &92.9$_{\pm1.0}$\%\\
\hline \makecell{NSCE=NCE+RCE} &94.6$_{\pm0.7}$\% &$95.5_{\pm0.5}$\%  &95.7$_{\pm1.0}$\%  &\textbf{91.1}$_{\pm1.2}$\% &\textbf{92.6}$_{\pm0.8}$\% &\textbf{93.1}$_{\pm1.0}$\%\\
\hline
\end{tabular}
\end{table}

The comparison presented in Tables  \ref{table:loss_comparison}, \ref{table:loss_comparison_swin} highlights the progressive improvement in Swin classification quality relative to ConvNext as the dataset size increases. It is worth noting that this phenomenon arises due to the well-established characteristic of transformers, which tend to exhibit higher efficacy with larger datasets while not being as efficient for smaller-scale datasets \cite{dosovitskiy2020image}.

\begin{table}[!t]
\caption{Swin classification metrics for different loss functions. 95\% confidence interval for the mean value}
\label{table:loss_comparison_swin}
\centering
\begin{tabular}{|c||c||c||c||c||c||c|}
\hline Model & \multicolumn{3}{c}{\thead{ROC AUC $\uparrow$ \\ frame \\ classification}} & \multicolumn{3}{c|}{\thead{Recall macro $\uparrow$ \\ CA view \\classification}} \\
\hline
&\thead{40\% data} & \thead{75\% data} & \thead{100\% data} &\thead{40\% data} &\thead{75\% data} &\thead{100\% data}\\
\hline \makecell{CE} &$92.5_{\pm0.8}$\% &$93.9_{\pm0.6}$\% &$94.9_{\pm0.8}$\% &$86.5_{\pm1.4}$\% &$89.7_{\pm1.0}$\% &$91.0_{\pm1.3}$\%\\
\hline \makecell{SCE=CE+RCE} &\textbf{93.0}$_{\pm0.8}$\% &\textbf{94.6}$_{\pm0.5}$\% &\textbf{95.1}$_{\pm0.7}$\% &90.1$_{\pm1.2}$\% &$91.5_{\pm0.8}$\% &92.7$_{\pm1.1}$\%\\
\hline \makecell{NSCE=NCE+RCE} &90.5$_{\pm1.0}$\% &$91.8_{\pm0.7}$\%  &93.1$_{\pm1.2}$\%  &\textbf{90.3}$_{\pm1.0}$\% &\textbf{92.2}$_{\pm0.8}$\% &\textbf{93.2}$_{\pm1.0}$\%\\
\hline
\end{tabular}
\end{table}

Label smoothing \cite{muller2019does} is a regularization technique that has been widely used in the field of deep learning to improve model generalizability. Our experiments, however, have shown that it only deteriorated the results of dominance classification when applied to our data set containing hard cases and false labels.

\section{Discussion}
Analyzing hard cases where the model failed can provide valuable insight into potential areas for further research into better solutions. To this end, additional data set labeling is necessary; specifically, labels such as RCA occlusions, and atypical branching should be included. Furthermore, to classify a heart in the case of RCA occlusions it is essential to utilize LCA information that could potentially influence dominance classification metrics - this issue requires further investigation. Additionally, uncertainty estimation and determination of vague cases are important tasks that must considered when conducting research on improving solutions for these difficult cases.

The future improvements of the model to our minds are:
\begin{itemize}
\item{ additional labelling of the data set with new tags such as bad quality, occlusion, atypical branching, and artefacts in order to better represent the data set;}
\item{utilizing LCA decision-making techniques for more accurate predictions}
\item{improving data set imbalance by adding new left dominant studies that can provide a better representation of all classes}
\item{predicting classification uncertainty, which will allow us to make more informed decisions}
\end{itemize}

\paragraph{Model explainability}
The neural network is a black box, and its logic is unclear, making the interpretation of results critical for clinical decision-making. To address this issue, we attempted to interpret the results by visualizing Swin attention layers and using Grad-CAM for ConvNext. Unfortunately, GradCam did not provide any clear interpretation in many cases; this result is consistent with study \cite{rodrigues2021automated}, which used GradCam for angiogram angle prediction interpretations. Meanwhile, GradCam showed better explainability in \cite{avram2021cathai} for stenosis classification.

\begin{figure*}[!t]
\centering
\subfloat[]{\includegraphics[width=1.2in]{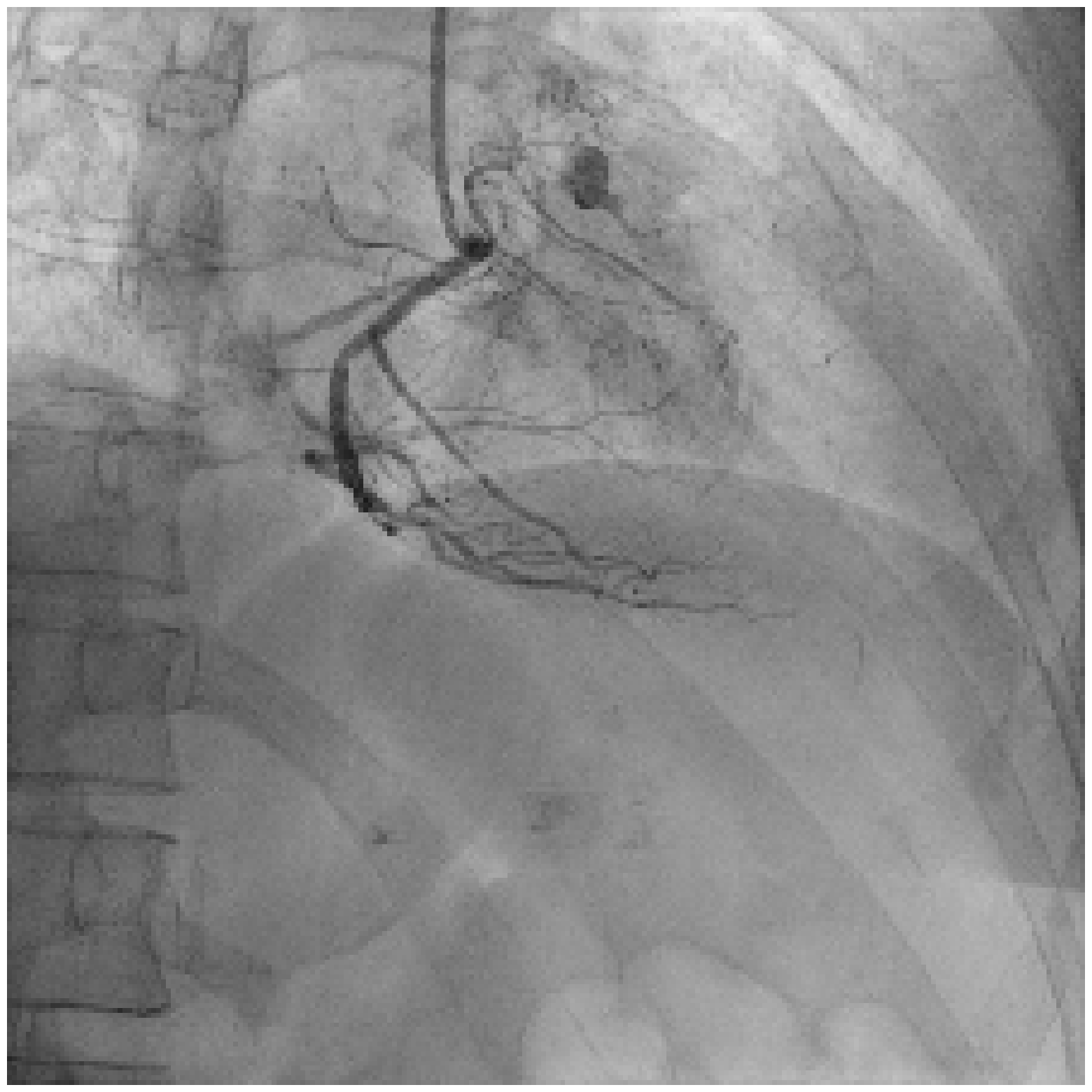}%
\label{fig_attention_orig}}
\hfil
\subfloat[]{\includegraphics[width=1.2in]{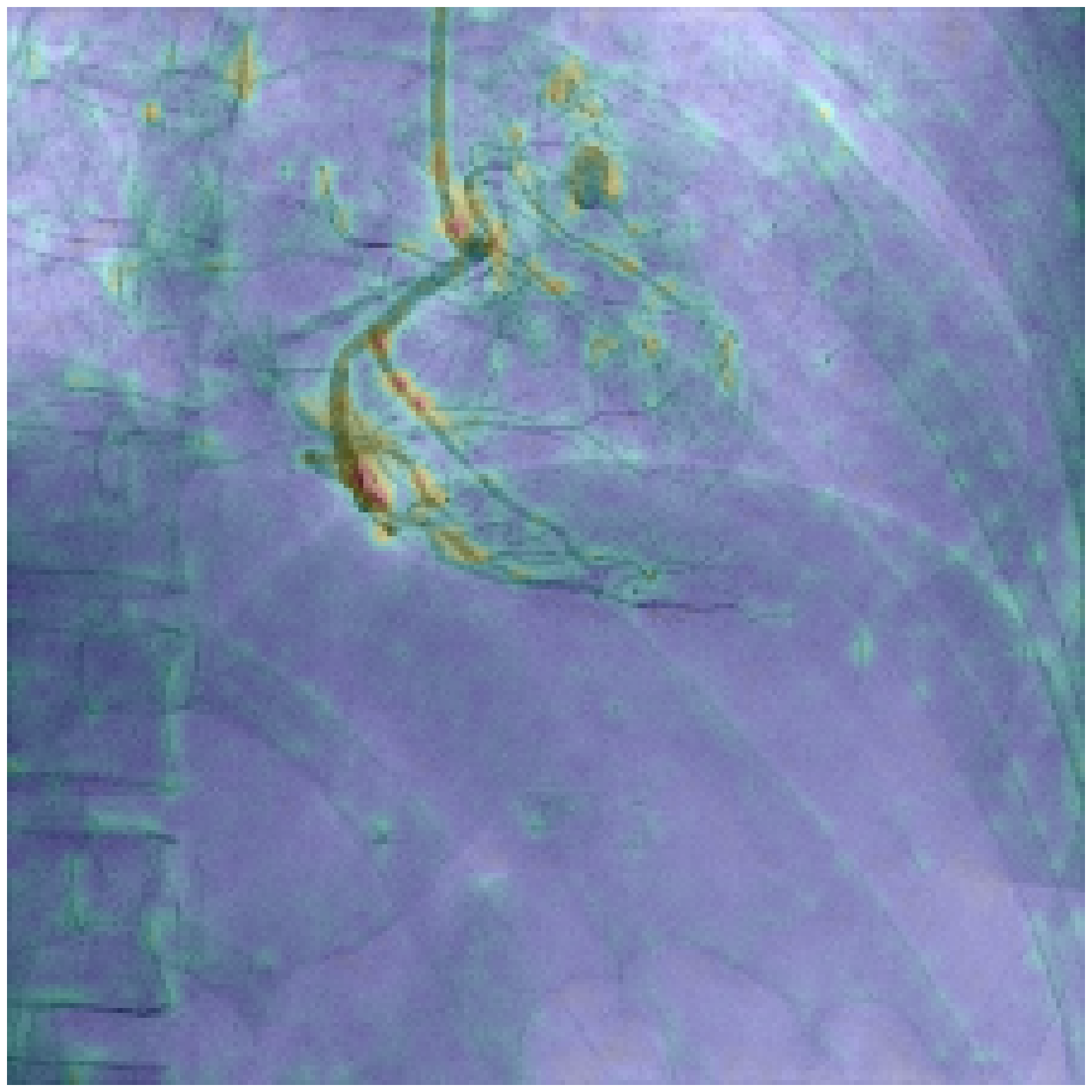}%
\label{fig_attention_1}}
\hfil
\subfloat[]{\includegraphics[width=1.2in]{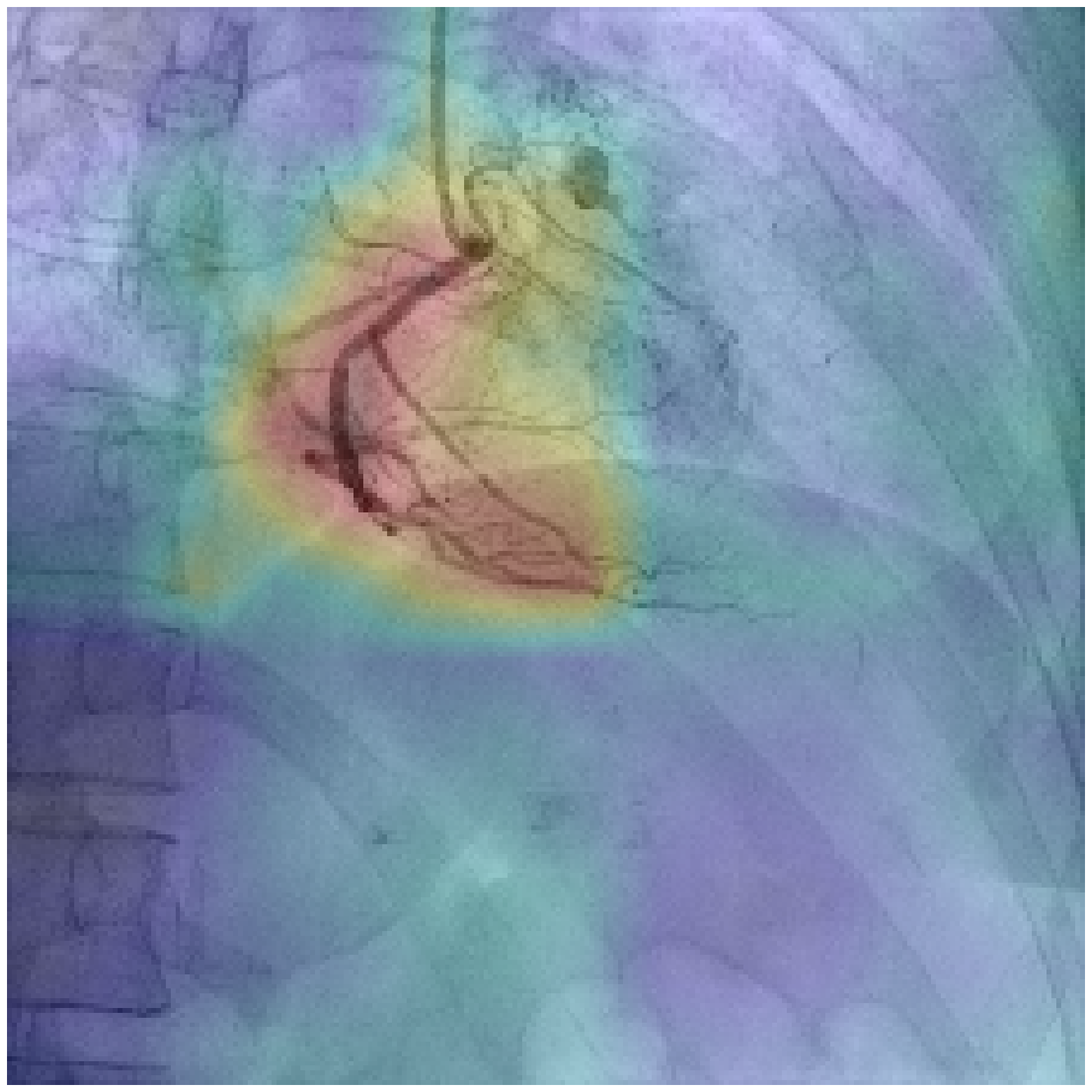}%
\label{fig_attention2}}
\subfloat[]{\includegraphics[width=1.2in]{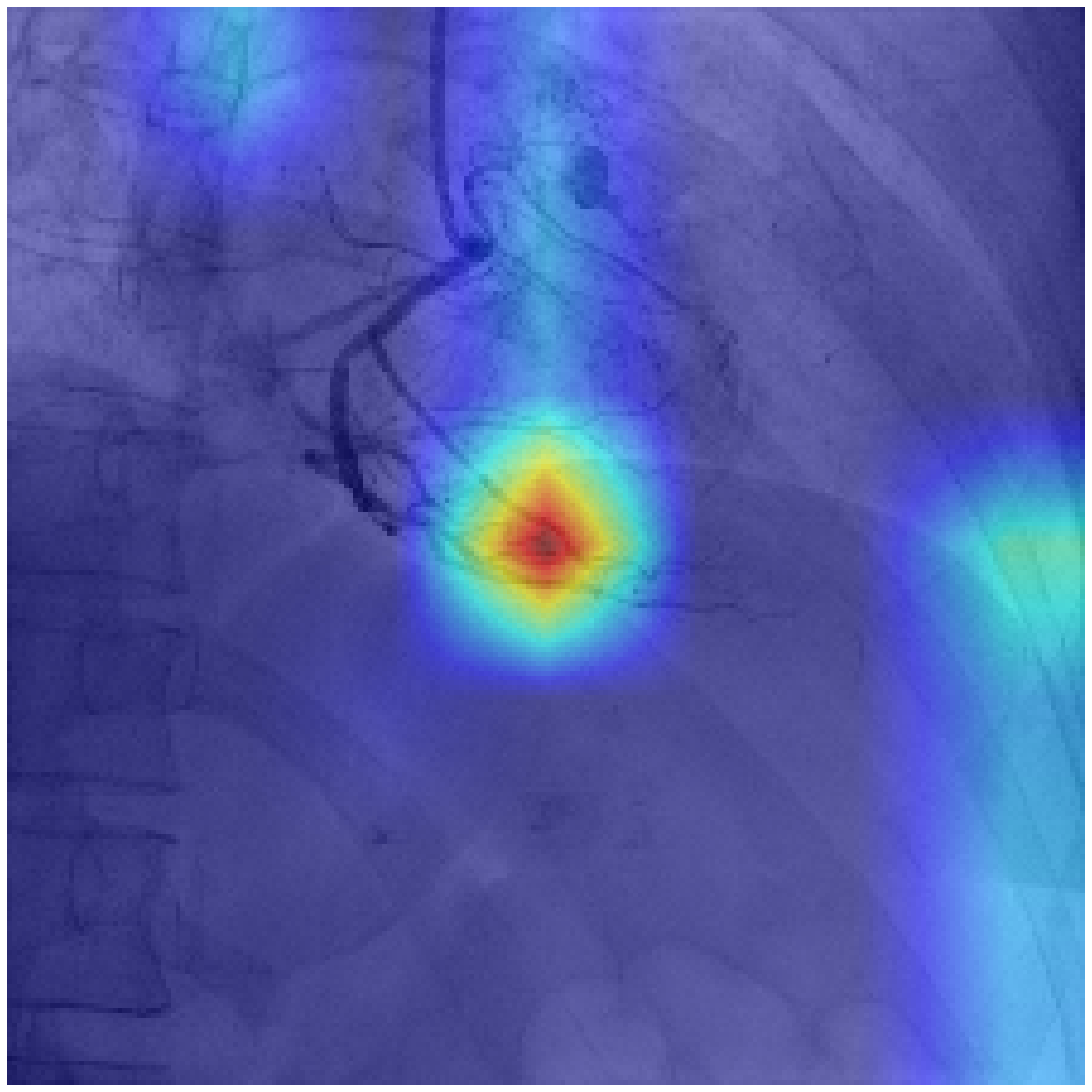}%
\label{fig_grad_cam}}
\caption{Swin attention visualization and GradCam for ConvNext (a) Original frame. (b) Attention layer of the 1st stage, 1st transformer block. (c) Attention layer of the 3d stage, 3d transformer block. (d) GradCam for ConvNext}
\label{fig_attention}
\end{figure*}

In contrast, visualization of Swin attention layers demonstrated better performance; as shown in Fig. \ref{fig_attention_1}, it focused on vessels and areas where branches are concentrated (Fig. \ref{fig_attention2}). Fig. \ref{fig_attention_1} highlights the result from window multi-head attention, wheres \ref{fig_attention2} shows the output from the end of the Swin transformer block after MLP layer.

Fig. \ref{fig_saturation} demonstrates that a model trained on a small data set of only 165 studies can achieve more than 90\% $recall_{macro}$, indicating the existence of simple criteria which allows for easy classification of the majority of cases. It is established that the right dominant heart vessels are typically thicker and have a larger square than the left. The model may use this simple criterion to classify hearts. However, our application of crop augmentation should prevent the model from relying on this simple criterion for classification.

Therefore, further research is needed to determine if models can analyze more complex structures such as RCA branches to achieve accurate results. Segmentation of all artery branches may be necessary to gain insight into this matter and develop better algorithms.

\paragraph{Model generalisability.}
Using two different models, a convolutional network and a transformer, combined with cross-validation has yielded consistent results. The five different models make mistakes mainly in the same studies with clear anatomical explanations. However, due to the relatively small size of our data set, there is a high deviation between folds caused by differences in train/val/test splits (the deviation between models trained and tested on the same data is small). Increasing the dataset size and mitigating any imbalance should help reduce this deviation between folders.

Our models were trained and tested on data from the same device, which does not prove their stability to domain shift. The indications for coronary angiography may differ between countries, leading to different data distributions that can cause a substantial increase in RCA occlusion cases. To ensure our models are robust across domains, we need to evaluate them using data sets from multiple sources with diverse characteristics.

\section{Conclusion}
The proposed machine learning approach to automatically classify cardiac dominance based on the RCA (Fig. \ref{fig_segment_scheme}) has been proven effective, with a classification accuracy of 93.5$\pm$3.8\%. The recall is comparable regardless of the type of dominance. However, in cases where an occlusion or atypical branching present in the RCA, it becomes necessary to utilize LCA information. These hard cases account for approximately 5\% of data sets; if excluded from consideration, this increases accuracy up to 97.0$\pm$2.1\%.

Both convolutional models ConvNext and Swin Transformer have shown comparable performance in the study. However, it is worth noting that Swin is biased toward left dominance, whereas ConvNext appears to be more balanced and has better metrics on the smaller data sets.

\bibliographystyle{splncs04}
\bibliography{sample}
%




\end{document}